%% file: ieee_tnnls.tex
\documentclass[journal, twoside]{IEEEtran}

\usepackage{times}
\usepackage{epsfig}
\usepackage{graphicx}
\usepackage{amsmath}
\usepackage{amssymb}
\usepackage[numbers,sort]{natbib}

\usepackage[utf8]{inputenc} 
\usepackage{amsfonts}       
\usepackage{nicefrac}       
\usepackage{microtype}      
\usepackage{multirow}
\usepackage{booktabs}
\usepackage{amsmath}
\usepackage{mathtools}
\usepackage{amssymb}
\usepackage{enumitem}
\usepackage{tabularx}
\usepackage{adjustbox}
\usepackage{threeparttable}
\usepackage{subfig}
\usepackage{algorithm}
\usepackage{algorithmic}
\usepackage{etoolbox}
\usepackage{lipsum}
\usepackage{cuted}
\usepackage{capt-of}
\usepackage{dsfont}

\usepackage[switch]{lineno}
\usepackage{color}
\usepackage{xcolor}
\usepackage{colortbl}
\usepackage{placeins}
\usepackage{kotex}
\usepackage{footnote}
\usepackage{bbding}
\usepackage{wrapfig}
\usepackage{tabulary}
\usepackage{relsize}
\usepackage[pagebackref=true,breaklinks=true,colorlinks,linkcolor={myred},citecolor={mygreen},bookmarks=false]{hyperref}

\usepackage{scalerel}
\usepackage{tikz}
\usetikzlibrary{svg.path}

\definecolor{orcidlogocol}{HTML}{A6CE39}
\tikzset{
  orcidlogo/.pic={
    \fill[orcidlogocol] svg{M256,128c0,70.7-57.3,128-128,128C57.3,256,0,198.7,0,128C0,57.3,57.3,0,128,0C198.7,0,256,57.3,256,128z};
    \fill[white] svg{M86.3,186.2H70.9V79.1h15.4v48.4V186.2z}
                 svg{M108.9,79.1h41.6c39.6,0,57,28.3,57,53.6c0,27.5-21.5,53.6-56.8,53.6h-41.8V79.1z M124.3,172.4h24.5c34.9,0,42.9-26.5,42.9-39.7c0-21.5-13.7-39.7-43.7-39.7h-23.7V172.4z}
                 svg{M88.7,56.8c0,5.5-4.5,10.1-10.1,10.1c-5.6,0-10.1-4.6-10.1-10.1c0-5.6,4.5-10.1,10.1-10.1C84.2,46.7,88.7,51.3,88.7,56.8z};
  }
}

\newcommand\orcidicon[1]{\href{https://orcid.org/#1}{\mbox{\scalerel*{
\begin{tikzpicture}[yscale=-1,transform shape]
\pic{orcidlogo};
\end{tikzpicture}
}{|}}}}

\newcommand{\figref}[1]{Fig.~\ref{#1}}
\newcommand{\tabref}[1]{Table~\ref{#1}}
\newcommand{\eqnref}[1]{Eq.~(\ref{#1})}
\newcommand{\secref}[1]{Sec.~\ref{#1}}
\newcommand{\eg}{\textit{e.g.}}
\newcommand{\ie}{\textit{i.e.}}
\newcommand{\vs}{\textit{vs.}}
\newcommand{\etal}{\textit{et al.}}

\usepackage[capitalize]{cleveref}
\crefname{section}{Sec.}{Secs.}
\Crefname{section}{Section}{Sections}
\Crefname{table}{Table}{Tables}
\crefname{table}{Tab.}{Tabs.}

\definecolor{todo}{rgb}{0.8, 0.4, 0.2}
\definecolor{Gray}{gray}{0.9}
\definecolor{myred}{rgb}{0.8, 0.1, 0.1}
\definecolor{mygreen}{rgb}{0.2, 0.7, 0.1}

\title{
    \textit{What and When to Look?}:\\
    Temporal Span Proposal Network for\\ Video Relation Detection
}

\author{
    Sangmin~Woo$^{\textsuperscript{\orcidicon{0000-0003-4451-9675}}}$,~\IEEEmembership{Student~Member,~IEEE,}
    Junhyug~Noh$^{\textsuperscript{\orcidicon{0000-0003-1239-8178}}}$,~\IEEEmembership{Member,~IEEE,}
    and~Kangil~Kim$^{\textsuperscript{\orcidicon{0000-0003-3220-6401}}}$,~\IEEEmembership{Member,~IEEE}
    \thanks{
    This work was supported by the National Research Foundation of Korea (NRF) grant funded by the Korea government (MSIT) (2019R1A2C109107712), and the Institute of Information \& communications Technology Planning \& Evaluation (IITP) grant funded by the Korea government (MSIT) (No. 2019-0-01842, Artificial Intelligence Graduate School Program (GIST)). \textit{(Corresponding author: Kangil Kim)}}
    \thanks{Sangmin Woo is with the School of Electrical Engineering, Korea Advanced Institute of Science and Technology, Daejeon 34141, Korea. This work was done when he was an M.S. student at GIST (email: smwoo95@kaist.ac.kr).}
    \thanks{Junhyug~Noh is with the Computational Engineering Division, Lawrence Livermore National Laboratory, CA 94550, United States (email: noh1@llnl.gov).}
    \thanks{Kangil Kim is with the School of Electrical Engineering and Computer Science and the AI Graduate School, Gwangju Institute of Science and Technology, Gwangju 61005, Korea (email: kangil.kim.01@gmail.com).}
}

\begin{document}
    \maketitle

    \begin{abstract}
        Identifying relations between objects is central to understanding the scene.
        While several works have been proposed for relation modeling in the image domain, there have been many constraints in the video domain due to challenging dynamics of spatio-temporal interactions (\eg, between which objects are there an interaction? when do relations start and end?).
        To date, two representative methods have been proposed to tackle Video Visual Relation Detection (VidVRD): segment-based and window-based.
        We first point out limitations of these methods and propose a novel approach named Temporal Span Proposal Network (TSPN).
        TSPN tells \textit{what to look}: it sparsifies relation search space by scoring relationness of object pair, \ie, measuring how probable a relation exist.
        TSPN tells \textit{when to look}: it simultaneously predicts start-end timestamps (\ie, temporal spans) and categories of the all possible relations by utilizing full video context.
        These two designs enable a win-win scenario: it accelerates training by 2$\times$ or more than existing methods and achieves competitive performance on two VidVRD benchmarks (ImageNet-VidVDR and VidOR).
        Moreover, comprehensive ablative experiments demonstrate the effectiveness of our approach.
        Codes are available at~\url{https://github.com/sangminwoo/Temporal-Span-Proposal-Network-VidVRD}.
    \end{abstract}
    
    
    \input{01introduction}
    \input{02related_work}
    \input{03method}
    \input{04experiments}
    \input{05conclusion}
    \input{08references}
    \input{09biography}

\end{document}

%% file: 01introduction.tex
\section{Introduction}
\label{sec:introduction}

\begin{figure}[t!]
    \centering
    \includegraphics[width=\linewidth]{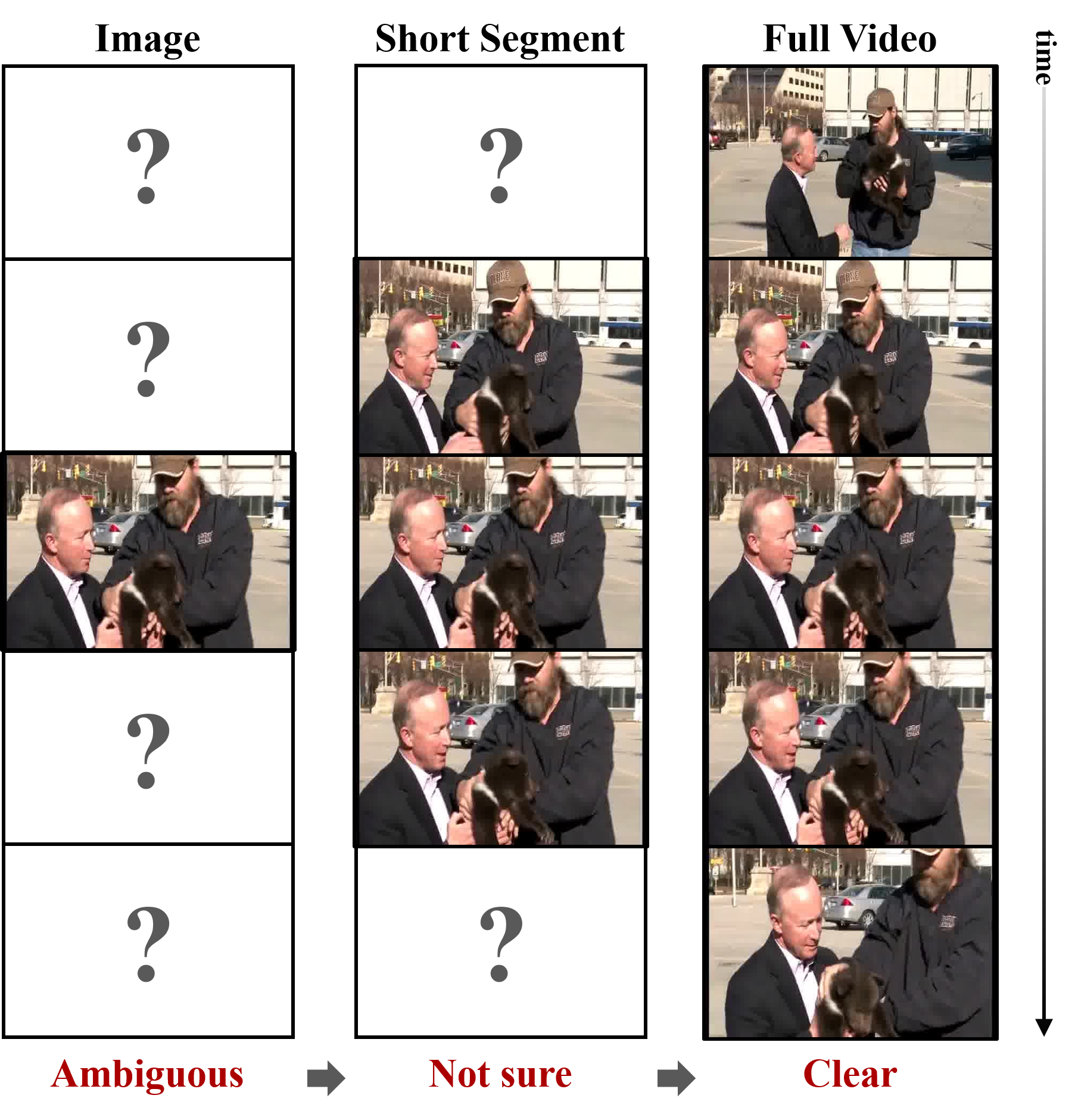}
    \caption{
        \textbf{Who is handing over the bear to whom?}
        When we try to answer the question with only a single image (leftmost), the answer can be both: left or right man.
        While guessing relations in short-term video segment (middle) is still questionable, the answer becomes clear in full video (rightmost) thanks to the spatio-temporal contexts.
        The time sequence is from top to bottom.
        \textbf{Answer: right man.}
    }
\label{fig:img_vs_vid}
\end{figure}

\IEEEPARstart{C}{apturing} the semantics of the visual scene has long been a concern in computer vision. Despite the remarkable progress of computer vision, understanding visual scenes remains a challenging task. 
On the way to leap forward, visual relations serve as the stepping stone for narrowing the gap between perceptive and cognitive tasks. A number of works leverages the visual relations including the tasks of image retrieval~\cite{johnson2015image}, dynamics prediction~\cite{santoro2017simple,battaglia2016interaction}, image captioning~\cite{yao2018exploring, yang2019auto}, visual question answering~\cite{teney2017graph}, image generation~\cite{johnson2018image,ashual2019specifying} and video understanding~\cite{ma2018attend}.
The Visual Relation Detection (VRD) task requires modeling both visible information about what and where entities are and underlying information of what interactions are happening between objects in the scene.
Relation reasoning is essential in the high-level understanding of the scene. 
Since a pioneer work of VRD~\cite{lu2016visual} was proposed, several interesting relational reasoning approaches have been studied on the image domain~\cite{dai2017detecting,zhang2017visual,jae2018tensorize,woo2022tackling}.
Recently, a Video Visual Relation Detection (VidVRD) task has been proposed~\cite{shang2017video}, but due to the difficulty of spatio-temporal relationship modeling, it has not yet been received much attention, and just a few of works~\cite{tsai2019video,qian2019video,sun2019video,su2020video,liu2020beyond,li2021interventional,shang2021video,chen2021social,zheng2022vrdformer} have been proposed.

\begin{figure*}[t!]
    \centering
    \includegraphics[width=\linewidth]{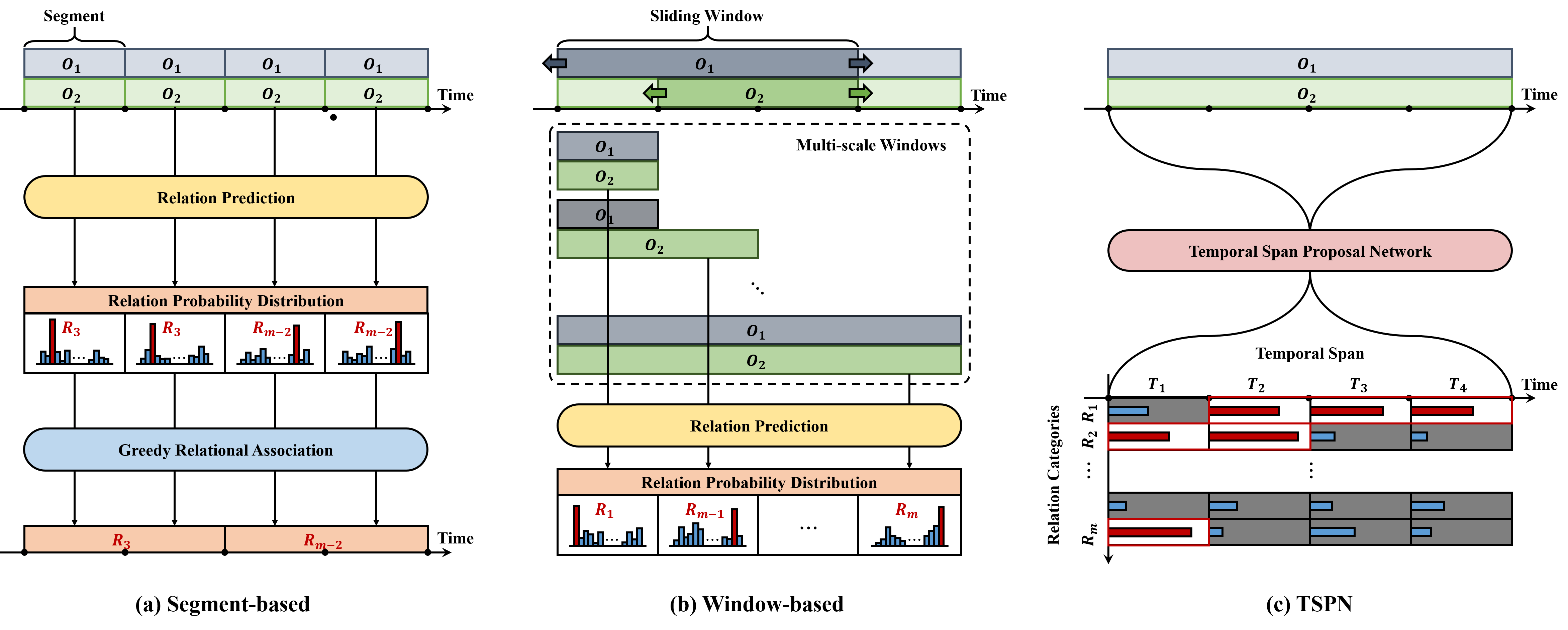}
    \caption{
        \textbf{Conceptual Comparison of typical VidVRD approaches} (empirical comparisons are in Table~\ref{tab:time}, \ref{tab:vidvrd}, \ref{tab:vidor}).
        \textbf{(a) Segment-based approach} first chunks a video into multiple segments, predict the short-term relations within each segment, and then greedily associate the relations of adjacent segments into the long-term relations.
        \textbf{(b) Window-based approach} generates a set of sub-tracklet pairs via a size-varying sliding window, and then predict all relations with different temporal span.
        \textbf{(c) TSPN (ours)} jointly predicts relation categories and its temporal span with a single video-level object trajectory pair.
        $O_i$ stands for $i$-th object trajectory of all object trajectories in the video, and $R_j$ denotes $j$-th relation category.
        We assume the relations are predicted only for temporal span in which two object trajectories appear simultaneously in the video. 
        We note that illustration of each method may not contain all the detailed procedures.
    }
\label{fig:comparison}
\end{figure*}

What makes detecting relations in video difficult? At first glance, we consider the common difficulties of the task.
In the view of the object, new objects may appear or disappear due to viewpoint variation or occlusion over time and may also contain motion blur.
Therefore, we need to adopt proper object detection~\cite{ren2015faster} and tracking~\cite{bewley2016simple,wojke2017simple} methods.
In the view of the relation, it requires modeling long-term temporal dependency and complex dynamics.
Relational reasoning in the video requires contextual information over time, unlike static images. For example, in~\figref{fig:img_vs_vid}, the relation between two men and bear is ambiguous and hard to define with a static image but can be inferred from longer spatio-temporal contexts.
Suppose there are $n$ object and $m$ relation categories, the number of possible combinations of pair-wise relationships is $O(n^2 \times m)$.
A naive approach to this problem is to learn a distribution over the $n \times n \times m$ lattice space -- the complete combinations of objects and relations. 
However, the model would be highly prone to bias in this case, due to the sparsity of existing relation types and data limits.
A simple sidestep is to learn low-rank matrices, factorized by object and relations.
To put it more simply, the strategy is to separate pipelines of predicting objects and relations, reducing the complexity to $O(n+m)$.
To this end, we first detect objects and then predict relations conditionally.
Here we further factorize the relation prediction process into two subprocesses: relationness scoring and temporal span proposal.
We detail the steps in~\secref{sec:tspn}.

To investigate the more complicated issues of the task, we compare typical VidVRD approaches in~\figref{fig:comparison}.
They can be divided into two categories: (a) segment-based and (b) window-based approach.
Segment-based approaches~\cite{shang2017video,di2019multiple,tsai2019video,qian2019video,sun2019video,su2020video} first split a video into several segments -- typically a segment contains 30 frames, and then predict relations at the segment-level.
If the neighboring segments share the same relation triplets, they are merged into a single relation with an extended range.
This approach predicts segment-wise relations with only corresponding 30 frames, assuming that the basic relations can always be found in a short duration and can indirectly build the video-level predictions via the association method.
However, we argue that 30 frames are not enough to model long-term interactions such as `\texttt{dog-past-person}'.
Because the relation `\texttt{past}' can only be inferred based on the overall process by which a dog is behind a person and overtakes a person, continuous monitoring of interactions between objects is required during successive events.
More recently, a window-based approaches~\cite{liu2020beyond,shang2021video,gao2021video} have been proposed in order to
incorporate longer temporal contexts of video.
They adopt a sliding window scheme over the whole object trajectories and obtains all possible sizes of object tracklets. 
With the object tracklets of various lengths, they can predict both short-term and long-term interactions.
However, we see that this is hardly scalable for extremely long videos such as movies because it requires an exhaustive search to sample all combinations of size-varying tracklet pairs to cover various durations of relations, resulting in a cubic complexity.

This work aims to bridge the gap between segment-based methods and window-based methods while improving the shortcomings and exploiting the advantages of the methods with proposed Temporal Span Proposal Network (TSPN) (see~\figref{fig:comparison}(c)).
Given a pair of object trajectories, TSPN first finds what pairs of objects are probable to have relations in-between\footnote{In this paper, we use the term `relationness' to indicate the probability that a relation exists between a pair of objects.}, then predicts when the pair-wise relations begin and end.
We call the first step as \textit{relationness scoring} and the second step as \textit{temporal span proposal}.
Since it is possible that an object pair can have multiple relations (\eg, A-standing next to-B; A-watch-B), we formalize the second step of TSPN as a multi-label classification in practice.
By leveraging both temporal locality as in the segment-based approach and temporal globality as in the window-based approach, TSPN can effectively predict pair-wise object relationships.
It is also clear that TSPN is more efficient than both methods because it only uses each object pair once to predict all relationships across the video.
In particular, it prevents duplicate use of each object tracklet by dividing it into different window lengths, such as the window-based method.
By design, the proposed TSPN can coarsen relation search space by capturing the regularity of pair-wise object interaction.
Also, TSPN leverages the global video context features, making it strong in both the short and long-term relationship modeling. 

We validate TSPN against existing approaches on two video visual relation detection datasets: ImageNet-VidVRD~\cite{shang2017video} and VidOR~\cite{shang2019annotating}.
We observe that our method achieves new state-of-the-art on both benchmarks with a significant performance gain.
Also, we examine the theoretical computation complexity of three methods (segment, window, and TSPN) and see that TSPN is approximately 2$\times$ and 4 orders of magnitude efficient than the segment-based method and window-based method, respectively.
Moreover, comprehensive quantitative and qualitative analyses demonstrate the efficacy of building blocks of TSPN.

Our contributions can be summarized as follows:
\begin{itemize}
\item We investigate the problem of Video Visual Relationship Detection (VidVRD), and identify the challenges that the task itself naturally entails and the challenges that the current (segment-based and window-based) methods have. To the best of our knowledge, TSPN is the first proposal-based approach for VidVRD.
\item We propose a novel approach named Temporal Span Proposal Network (TSPN) that tackles the aforementioned issues. TSPN has two key components: (i) relationness scoring module that sparsifies relation search space and (ii) temporal span proposal module that predicts relations and its duration upon global video context.
\item We show that TSPN is not only effective but also efficient: TSPN achieves state-of-the-art performances in two benchmark datasets (ImageNet-VidVRD and VidOR) and theoretically faster than conventional VidVRD methods by 2$\times$ or more.
\item Comprehensive ablation studies demonstrates the efficacy of model components and several design choices.
\end{itemize}

%% file: 02related_work.tex
\section{Related Work}
\label{sec:related_work}
Our work intersects the following three research topics: multi object tracking, relation detection in images \& videos, and proposal networks. 

\subsection{Multi Object Tracking}
Multiple Object Tracking (MOT) aims to identify and track objects in videos without prior knowledge of the appearance and number of targets.
Unlike object detection algorithms~\cite{ren2015faster,redmon2016you,liu2016ssd}, where the output is a set of rectangular bounding boxes with four coordinates (or two coordinates with height and width), MOT methods~\cite{bewley2016simple,wojke2017simple,zhu2018online,fu2018gm,sheng2018heterogeneous,kim2018multi} assign an additional target ID to each box in order to differentiate objects within the same class.
Standard MOT algorithms follow a tracking-by-detection strategy.
First, a collection of bounding boxes are extracted  from the video frames (detection), and then they are utilized to guide the tracking process (tracking).
The tracking is typically done by bounding box association across the frames: bounding boxes containing the same target are assigned with the same ID.


As VidVRD operates upon object trajectory proposals, we employ the tracking-by-detection strategy for MOT.
In line with other VidVRD works~\cite{shang2017video,liu2020beyond}, we utilize Faster R-CNN~\cite{ren2015faster} for object detection and DeepSORT~\cite{wojke2017simple} for multi object tracking from detected bounding boxes.

\subsection{Relation Detection in Images}
In an effort to detect relations in images, numerous studies have been explicitly modeled and adopted deep neural networks.
The challenging and open-ended nature of the task lends itself to various forms.

Visual Relation Detection (VRD)~\cite{lu2016visual,zhang2017visual,liang2017deep,yu2017visual,dai2017detecting,zhuang2017towards} aims to recognize relations between objects in an image, which is the generic form of visual relational reasoning tasks.
Lu \etal~\cite{lu2016visual} present the first VRD work, in which they employed RCNN~\cite{girshick2014rich} to recognize objects and predicates in an image, and leverage language priors from semantic word embeddings in order to finetune the likelihood of predicted relationships.
Since then, a variety of methods has been proposed: visual translation embedding (VTransE)~\cite{zhang2017visual}; reinforcement learning (RL)-based framework~\cite{liang2017deep}; linguistic knowledge distillation~\cite{yu2017visual}; prior and posterior statistics~\cite{dai2017detecting}; context-aware attention models~\cite{zhuang2017towards}.

Scene Graph Generation (SGG)~\cite{xu2017scene,yang2018graph,zellers2018neural,woo2018linknet,tang2019learning,tang2020unbiased,woo2022tackling} is a graphical form of VRD, or a structured scene descriptor, where nodes correspond to objects and edges correspond to pairwise relationships between objects.
The challenges in SGG yield diverse methods : Xu \etal~\cite{xu2017scene} design an RNN-based model to exploit contextual cues in order to improve the scene graph predictions via iterative message passing between objects and predicates; Yang \etal~\cite{yang2018graph} utilizes graph convolution network~\cite{kipf2017semi} to model contextual information between objects and relations; Zellers \etal~\cite{zellers2018neural} analyzes the role of motifs: regularly appearing substructures in scene graphs and provide RNN-based model that captures higher order motifs; Tang \etal~\cite{tang2020unbiased} present unbiased SGG framework that is robust to skewed data distribution using causal inference; Woo \etal~\cite{woo2022tackling} identifies underlying challenges in SGG (ambiguity, asymmetry, higher-order contexts) and tackle the problem with local-to-global interaction network.

Human-Object Interaction (HOI)~\cite{chao2018learning,gkioxari2018detecting,li2019transferable,qi2018learning,zhou2021cascaded} more focuses on human and considers them as the subject.
Gkioxario \etal~\cite{dai2017detecting} build their model on Faster R-CNN~\cite{ren2015faster} with a human-centric branch that performs target object localization and action classification, and an interaction branch that combines human features with target object features; Qi \etal~\cite{qi2018learning} apply a graph neural network (GNN)-based model on the human-object graph containing all potential human-object interactions to remove unlikely edges to be connected; Zhou \etal~\cite{zhou2021cascaded} propose a multi-stream network which incorporates language priors, geometric features, and visual features.

\subsection{Relation Detection in Videos}
Most of the relation detection works are mainly conducted in the static image domain.
As a breakthrough, the pioneer work~\cite{shang2017video} has introduced the first dataset and the baseline of VRD in the dynamic video domain.
Since then, a few works have been proposed due to the difficulty of spatio-temporal modeling.

A majority of existing methods follow the segment-based approach~\cite{di2019multiple,qian2019video,tsai2019video,sun2019video,su2020video,xie2020video,li2021interventional,cao2021vsrn} which was firstly introduced in~\cite{shang2017video}.
It first breaks a video into several segments, predicts per-segment relations, and finally associates relations in a greedy manner.
This line of methods achieved sound performance, but they are inherently incapable of using the long-term temporal context in other segments: VidVRD~\cite{shang2017video} detects relations in the short-term and greedily associates them; MHRA~\cite{di2019multiple} generates multiple hypotheses for video relation instances for more robust long-term relation prediction; GSTEG~\cite{tsai2019video} constructs a conditional random field on a spatio-temporal graph exploiting the statistical dependency of relational entities; VRD-GCN~\cite{qian2019video} passes the message through fully-connected spatial-temporal graphs and conducts reasoning in the 3D graphs using Graph Convolution Network (GCN)~\cite{kipf2017semi}; MMFF~\cite{sun2019video} predicts relations by jointly using spatial-temporal visual feature and language context feature; MHA~\cite{su2020video} maintains multiple possible relation hypotheses during the association process to handle the inaccuracy of the former steps; VSRN~\cite{cao2021vsrn} utilizes 3D CNN to encode spatio-temporal information, and to use semantic collocations between objects for comprehensive relation representations; IVRD~\cite{li2021interventional} adopts a causal intervention on the input subject and object that leads the model to incorporate each potential predicate prototype, which is a set of relation references with the same predicate.

More recently, lie \etal~\cite{liu2020beyond} proposed a window-based approach to handle long-term temporal context.
It utilizes sliding window of varying sizes to generate a large number of tracklets then predicts their pairwise relations.
Since then, a handful of works have been proposed that follow window-based approach~\cite{shang2021video,gao2021video}: PPN-STGCN~\cite{liu2020beyond} employs a sliding-window scheme to predict both short-term and long-term relationships and utilizes GCN to calculate the compatibility of tracklet proposal pair; Gao \etal~\cite{gao2021video} construct a VidVRD model based on the Transformer encoder-decoder~\cite{vaswani2017attention} that models interactions between tracklets and learnable predicate queries; VidVRD-II~\cite{shang2021video} leverages inter-dependency among subject-object-predicate classification and fine-tunes its predictions based on the learnt dependency.



However, the above-mentioned methods either lack temporal globality or scale poorly for extremely long videos such as movies.
To move one step forward, we introduce a novel TSPN to directly predict relations in an end-to-end manner without the need for many heuristics such as segment length and window size.

\subsection{Proposal Networks} 
There is a large body of literature on object proposal techniques.
Standard object proposal approaches fall into two categories: 
(i) unsupervised approaches using super-pixel grouping~\cite{uijlings2013selective,arbelaez2014multiscale} or sliding windows~\cite{alexe2012measuring,zitnick2014edge}.
These techniques are mostly offline algorithms that operate independently of detectors (\eg, RCNN~\cite{girshick2014rich}, and Fast R-CNN~\cite{girshick2015fast}).
(ii) supervised approaches based on learnt deep representations from CNNs~\cite{ren2015faster,liu2016ssd,woo2018stairnet}.
In these approaches, proposal generation is done by a single forward pass inside an end-to-end trainable detection system.

Among them, Region Proposal Network (RPN)~\cite{ren2015faster} used for the object detection strongly inspired our TSPN.
RPN simultaneously predicts object bounding boxes and `objectness' score (we named the term `relationness' after this) to detect the region of interest.
We extend the concept of proposal from the object to relation as well as the space domain to the time domain.

Several works also share a similar spirit with TSPN that of reducing relation search space~\cite{zhang2017relationship,yang2018graph,liu2020beyond}:
Zhang \etal~\cite{zhang2017relationship} propose a Relation Proposal Network that selects a set of subject-object pairs among every possible pairs by evaluating visual and spatial compatibility; Yang \etal~\cite{yang2018graph} present a network that learns to estimate the relatedness of an object pair and prunes unlikely relations while preserving likely ones; Liu \etal~\cite{liu2020beyond} construct two graphs to independently formulate spatial and temporal interactions between tracklet proposal features to filter out incompatible proposal pairs.


%% file: 03method.tex
\section{Temporal Span Proposal Network (TSPN)}
\label{sec:tspn}
\begin{figure*}[t!]
    \centering
    \includegraphics[width=\linewidth]{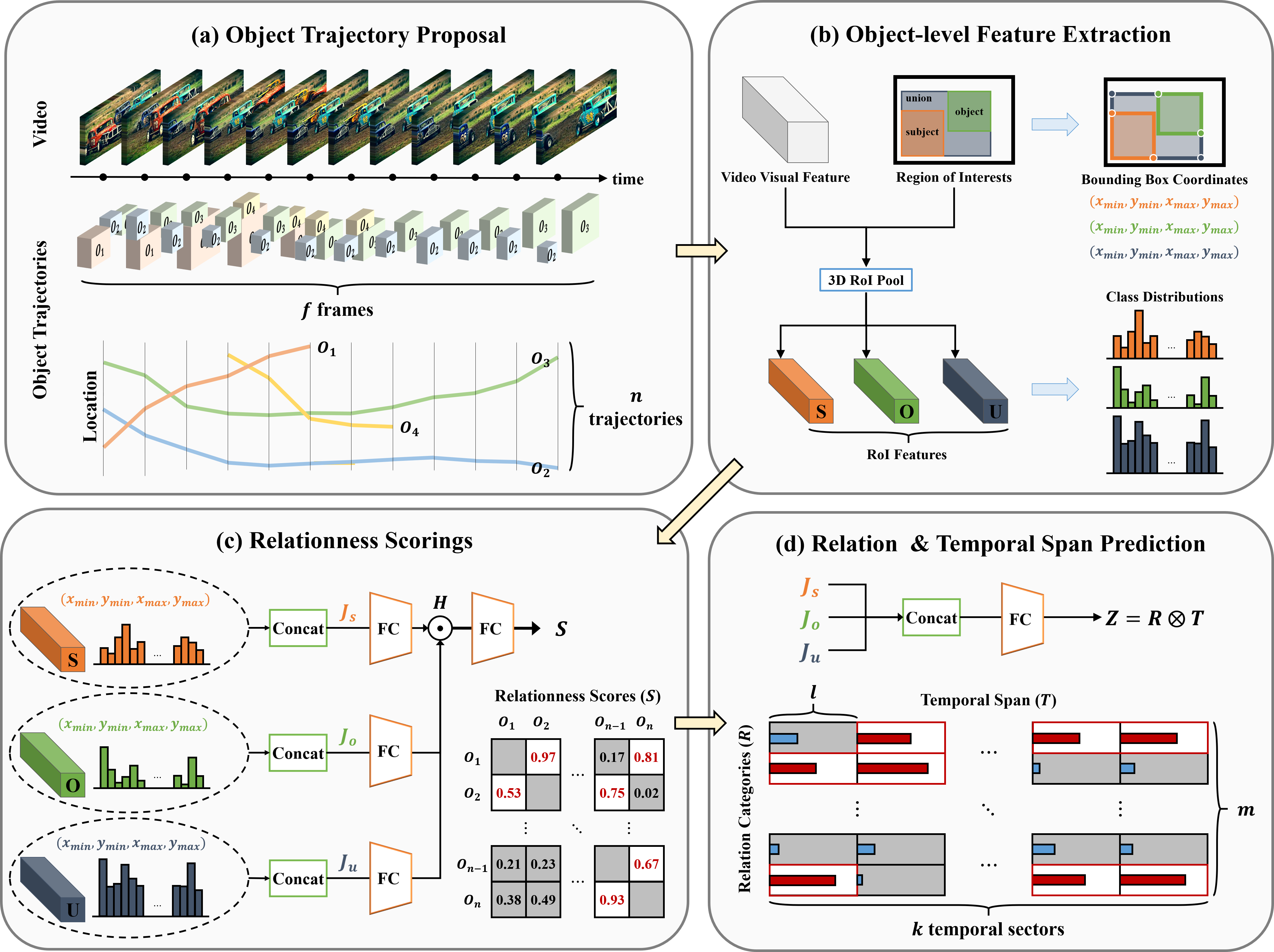}
    \caption{
        \textbf{Overview of TSPN.} 
        \textbf{(a)} TSPN is built upon the object trajectory proposal head which comprises object detection and tracking stages (\secref{sec:3b}).
        \textbf{(b)} We first extract video visual features via a CNN backbone~\cite{he2016deep}.
        With detection results (RoIs), we then extract RoI features~\cite{xu2017r} of subject, object, and union area.
        Their corresponding bounding box coordinates and class distribution can be naturally obtained from object detection phase (\secref{sec:3c}).
        \textbf{(c)} The concatenation of RoI features with bounding box coordinates and class distribution ($J$) are linearly transformed and then fused via a Hadamard product (denoted as $\odot$ in the figure), resulting in $H$. 
        The relationness $\mathcal{S}$ between an object pair is first calculated by feeding $H$ into a FC layer.
        After then, a set of pairs with high relationness scores (colored in red in the figure) is only considered in the subsequent process.
        Note that the relationness score is computed differently for $\mathcal{S}(O_1 \rightarrow O_2)=0.97$ and $\mathcal{S}(O_2 \rightarrow O_1)=0.53$ since the pair-wise relationship can vary when subject and object are switched (\secref{sec:3d}).
        \textbf{(d)} Finally, joint features are concatenated and fed to another FC layer to predict output $Z$ which is deemed as an outer product of relation labels $\mathcal{R}$ and their temporal spans $\mathcal{T}$, \ie, start-end time (\secref{sec:3e}).
        Our TSPN can be trained in an end-to-end manner.
        See texts for more details.
    }
    \vspace{-1mm}
\label{fig:overview}
\end{figure*}

In this section, we present an end-to-end trainable model Temporal Span Proposal Network (TSPN) for VidVRD.
TSPN is built on top of the object trajectory proposal module, which is a combination of object detector~\cite{ren2015faster} and multi-object tracker~\cite{wojke2017simple}.
In order to ease the optimization, we pre-train object detector and jointly train the whole video relation detection network.
TSPN shares the video-level convolutional features, enabling itself to make cost-efficient proposals.
An overview of TSPN is shown in~\figref{fig:overview}, and~\tabref{tab:notations} summarizes the relevant notations.

\begin{table}[h!]
    \setlength{\tabcolsep}{5pt}
    \centering
    \scriptsize
    \resizebox{\linewidth}{!}{
    \begin{tabular}{@{}ccc@{}}
        \toprule
        \textbf{Step} & \textbf{Notation} & \textbf{Description}\\ \midrule \midrule
        \multirow{4}{*}{\begin{tabular}[c]{@{}c@{}}Object Trajectory\\ Proposal\end{tabular}}
        & $\mathcal{V}$ & input \textbf{v}ideo \\
        & $\mathcal{O}$ & \textbf{o}bject trajectories \\
        & $f$ & number of \textbf{f}rames in input video \\
        & $n$ & \textbf{n}umber of object trajectories \\ \midrule
        
        \multirow{3}{*}{\begin{tabular}[c]{@{}c@{}}Object-level\\ Feature Extraction\end{tabular}}
        & $A$ & RoI \textbf{a}ppearance features \\
        & $B$ & bounding \textbf{b}ox coordinates \\ 
        & $C$ & \textbf{c}lassification probability distributions \\ \midrule
        
        \multirow{3}{*}{\begin{tabular}[c]{@{}c@{}}Relationness\\ Scoring\end{tabular}}
        & $J$ & \textbf{j}oint features \\
        & $\mathcal{S}$ & relationness \textbf{s}cores \\
        & $p$ & number of \textbf{p}air of interests \\ \midrule
        
        \multirow{6}{*}{\begin{tabular}[c]{@{}c@{}}Relation \&\\ Temporal Span Prediction\end{tabular}}
        & $\mathcal{R}$ & \textbf{r}elation categories \\
        & $\mathcal{T}$ & \textbf{t}emporal spans \\ 
        & $Z$ & outer product of $\mathcal{R}$ and $\mathcal{T}$ \\
        & $m$ & number of relation categories \\
        & $k$ & number of temporal sectors \\
        & $l$ & \textbf{l}ength of each sector \\ 
         \bottomrule
    \end{tabular}
    }
    \caption{\textbf{Summary of notations.}}
    \label{tab:notations}
\end{table}

\subsection{Problem Formulation}
We design a video relation detector with the notion of object trajectories, relationness, temporal span, and relation labels.
Formally, let $\mathcal{V}$ denote an input video, $\mathcal{O}$ be a set of object trajectories and $\mathcal{S}$, $\mathcal{T}$ and $\mathcal{R}$ denote the relationness score, temporal span, and relation, respectively.
The goal is to build a model for $P({\theta}_{\text{VidVRD}} = (\mathcal{O}, \mathcal{S}, \mathcal{T}, \mathcal{R}) | \mathcal{V})$.
The VidVRD can be factorized into three processes:
\begin{eqnarray}
    \begin{split}
        P(&{\theta}_{\text{VidVRD}}|\mathcal{V}) = \\
        &\underbrace{P(\mathcal{O}|\mathcal{V})}_{\substack{\text{Object Trajectory}\\ \text{Proposal}}}\underbrace{P(\mathcal{S}|\mathcal{O},\mathcal{V})}_{\substack{\text{Relationness} \\ \text{Scoring}}} \underbrace{P(\mathcal{T}, \mathcal{R}|\mathcal{S}, \mathcal{O}, \mathcal{V}).}_{\substack{\text{Relation \&} \\ \text{Temporal Span Prediction}}}
    \end{split}
\label{eq:factorization}
\end{eqnarray}
The most fundamental yet essential part of relation reasoning is the reliable object trajectory generation, since relationships are established between two objects.
Thus, we first detect locations and predict categories of objects appearing within the video.
The object trajectory proposal $P(\mathcal{O}|\mathcal{V})$ is typically modeled using an off-the-shelf object detector~\cite{ren2015faster} and multi-object tracker~\cite{wojke2017simple}.
The subsequent steps can be break down into 1) relationness scoring and 2) relation \& temporal span prediction subprocesses.
Given a set of object trajectory pairs, TSPN first calculates the probability that each pair will have a relation (\ie, relationness), and sample pairs of object trajectories with high relationness (\ie, pair-of-interest).
We note that even for the same pair of objects, relationness can be different when the semantic identity of the subject and the object is reversed: $\mathcal{S}(\mathcal{O}_1 \rightarrow \mathcal{O}_2) \neq \mathcal{S}(\mathcal{O}_2 \rightarrow \mathcal{O}_1)$.
TSPN finally predicts when relations start and end (\ie, temporal span) for all relation categories within a pair-of-interest.
Note that a pair-of-interest can have multiple relations: $\mathcal{O}_1 \xrightarrow{\mathcal{R}_1, \cdots , \mathcal{R}_m} \mathcal{O}_2$.

\subsection{Object Trajectory Proposal}
\label{sec:3b}
As all the subsequent processes heavily rely on the object trajectory proposals, it is essential to assure them to be high-quality.
The purpose of this step is to find trajectories $\mathcal{O}$ for all $n$ objects present in the video $\mathcal{V}$ with $f$ frames.
We adopt the Faster R-CNN~\cite{ren2015faster} object detector equipped with ResNet-101 backbone~\cite{he2016deep} to detect what the objects are (classification) and where they are located (localization) at every frame.
The Non-Maximum Suppression (NMS) algorithm~\cite{han2016seq} is performed with Intersection over Union (IoU) threshold of 0.5 to reduce redundant bounding boxes.
The object bounding boxes have variations in scale, aspect ratio, and position, hence we use Region-of-Interest (RoI) Align operation~\cite{he2017mask} to generate fixed-length feature representations from variable-size bounding box proposals, easing the subsequent relation computations.
From object detection, we can obtain object bounding boxes $B_1, \cdots, B_n$, which are defined by two points on the lower left and upper right (${x}_{min}$, ${y}_{min}$, ${x}_{max}$, ${y}_{max}$), and classification probability distributions $C_1, \cdots, C_n$, which are predicted from RoI appearance features $A_1, \cdots, A_n$ for each frame.

To ease the optimization of TSPN, we pre-train the object detector and then jointly train the whole TSPN model (including the object detector).
Let the number of objects in a video as $n$ and the possible number of relations as $m$.
Considering all pair-wise relations of $n$ objects, direct optimization of object detector and relation detector costs $O(n^2 \times m)$, which is prohibitively expensive; however, the sequential optimization of object and relation detectors lowers the costs to $O(n+m)$.

We also utilize a DeepSORT tracker~\cite{wojke2017simple} to link frame-level bounding boxes with the same identity into the video-level object trajectory proposals $\mathcal{O}_{1}, \cdots , \mathcal{O}_{n}$.
The DeepSORT tracker integrates visual features as matching descriptors to improve tracking performance.
We use RoI appearance features obtained from the object detection phase as visual features for DeepSORT.

\subsection{Object-level Feature Extraction}
\label{sec:3c}
For a pair of objects $\mathcal{O}_{s}, \mathcal{O}_{o}$, we obtain a set of RoI appearance features $A_s$, $A_o$, $A_u$ by applying RoI Align operation, where subscript $s$ and $o$ denote the semantic identity (\eg, subject and object) of objects, and $A_u$ denotes the union RoI feature.
The object class distributions $C_s$, $C_o$, $C_u$ are predicted via a linear predictor, \eg, FC layer, using RoI features $A_s$, $A_o$, $A_u$, where $C_u$ is defined as the mean of $C_s$ and $C_o$.
We define the union RoI as a rectangular area that tightly encompasses the subject RoI and object RoI (see~\figref{fig:overview} (b) for better understanding).
The frame-wise features $A, B, C$ are averaged across the frames.
We use these obtained features for the subsequent processes.


\subsection{Relationness Scoring}
\label{sec:3d}
The computational cost to learn relation distributions for every object pair is prohibitively expensive.
Only a handful of objects have meaningful relations due to the sparse nature of real-world interactions.
In order to model such regularities, we start by scoring relationness between object trajectory proposals.
More formally, from $n$ object trajectory proposals $\mathcal{O}$, the possible pairs are in ${n}^{2} \choose 2$.
Our goal here is to sample $p$ $\ll {n}^{2}$ pairs.

To this end, we filter out less correlated pairs based on the relationness score. We first jointly model visual, geometric and semantic relationness by concatenating RoI appearance features $A_i$ along with corresponding bounding box coordinates $B_i$ and classification probability distributions $C_i$, resulting in joint representation $J_i$.
\begin{eqnarray}
    {J}_{i} = {A}_{i} \parallel {B}_{i} \parallel {C}_{i},\quad \text{where,\ \ } i \in \{\textit{s,o,u}\}
    \label{eq:concatenate}
\end{eqnarray}
and $\parallel$ is a concatenation operation.

Then, each $J_i$ is projected into its respective embedding space via the FC layer, and fused by the Hadamard product (\ie, element-wise multiplication).
\begin{eqnarray}
    H = (\textbf{W}_{s}^{\textrm{T}} {J}_{s} + \textbf{b}_{s}) \odot (\textbf{W}_{o}^{\textrm{T}} {J}_{o} + \textbf{b}_{o}) \odot (\textbf{W}_{u}^{\textrm{T}} {J}_{u} + \textbf{b}_{u}),
    \label{eq:Z}
\end{eqnarray}
where $\textbf{W}$ and $\textbf{b}$ respectively denotes weight matrix and bias matrix of each FC layer (subscripts $s$, $o$, $u$ denote subject, object, and union, respectively).
The Hadamard product has shown to be effective as an attention mechanism for high-dimensional representations~\cite{kim2016hadamard}.
Also, unlike concatenation that increases the input dimension of the fusion unit (FC layer), the Hadamard product is efficient fusion operation since it maintains the same dimension.
The fused representation $H$ is passed to another FC layer to project them into the same space where the relationness score $\mathcal{S}$ is calculated.
Formally, the relationness score $\mathcal{S}$ is computed as follows:
\begin{eqnarray}
    \mathcal{S} = \sigma \left(\textbf{W}^{\textrm{T}} H + \textbf{b}\right),
\label{eq:relationness}
\end{eqnarray}
where $\textbf{W}$ and $\textbf{b}$ are weight matrix and bias matrix of FC layer, $\odot$ indicates Hadamard product, and $\sigma$ is a non-linear activation, which maps any real values into a finite interval (\eg, sigmoid).
For a pair of objects, subject ($s$) and object ($o$) can be switched. As such, the relationness score is calculated twice since it can vary depending on direction ($\mathcal{S}(\mathcal{O}_s \rightarrow \mathcal{O}_o) \neq \mathcal{S}(\mathcal{O}_o \rightarrow \mathcal{O}_s$)).

After iterating the process over every object trajectory pairs existing in the video\footnote{In practice, $n$ object trajectories are batched together to calculate pair-wise relationness scores in parallel.}, we sort the output relationness scores in descending order and maintain only the top-$p$ pair-of-interests per video.
Only these are considered in subsequent processes.

\begin{table*}[t!]
    \setlength{\tabcolsep}{8pt}
    \centering
    \resizebox{0.8\linewidth}{!}{
    \begin{tabular}{@{}ccccc@{}}
        \toprule
        Method & Approximated Time & Upper Bound & Typical Setting ($s=\frac{l}{2}$) & Big-$O$ \\ \midrule
        Segment-based & $\left\lceil \frac{L-l+s}{s} \right\rceil$ & $\frac{L-l+2s}{s}$ & $\frac{2L}{l}$ & $O(L)$ \\
        Window-based  & $\sum_{k=1}^{\frac{L}{l}} \left\lceil \frac{L-kl+s}{s} \right\rceil$ & $\left( \frac{L^2}{s} - \frac{L^2}{2l^2s} - \frac{L}{2ls} + 2L \right)^2$ & $\left( \frac{2L^2}{l} - \frac{L^2}{l^3} - \frac{L}{l^2} + 2L \right)^2$ & $O(L^4)$ \\ \midrule
        TSPN (ours) & $\frac{L}{l}$ & $\frac{L}{l}$ & $\frac{L}{l}$ & $O(L)$ \\ \bottomrule
    \end{tabular}
    }
    \caption{
        \textbf{Comparison of estimated relation prediction time with typical approaches.}
        $L$ denotes the intersection of subject trajectory and object trajectory (\ie, pair-of-interest), $l$ denotes segment length (in segment-based) or minimum window size (in window-based) or coverage of each temporal sector (in TSPN), and $s$ denotes the stride of segment (in segment-based) or window (in window-based), where $s < l \ll L$.
        Note that $l$ is set the same for all methods for a more equitable comparison.
        We conclude that our TSPN is 2$\times$ or more faster than conventional VidVRD methods in typical setting, where $s=\frac{l}{2}$.
    }
    \label{tab:time}
\end{table*}

\subsection{Relation \& Temporal Span Prediction}
\label{sec:3e}
Due to the dynamic nature of video, numerous relations may exist between a single pair of object trajectories.
These relations can be exist in the same time span or can be placed over multiple time spans.
For example, think of a scenario where two racehorses appear: two racehorses stand side by side, move ahead when they hear a gunshot, and continue to compete during the race.
Even to simplify the situation, it would be \texttt{horse1} -- \texttt{stand next to} -- \texttt{horse2} in the early timespan, and \texttt{horse1} -- \texttt{chase} -- \texttt{horse2} in the late timespan.
Therefore, we need to simultaneously predict what relations exist and when the relations occur (\ie, the range of time), given the object trajectory pair.

There is an open choice in predicting the temporal span of relations.
Early object detectors adopt RPN~\cite{ren2015faster} that uses a sliding window scheme with anchors in different scales to learn region proposals by reducing the localization error between anchors and actual object regions.
However, when it comes to temporal span proposal, it becomes unnecessary since relation detection does not require fine-grained temporal duration to match the ground-truth.
We thus are not using the anchor or sliding-window scheme in this work.
The brute-force approach to find temporal span is to learn a probability distribution over the ${m}\times{f}$ lattice space -- the number of relation categories is $m$, and the total number of frames is $f$ -- which is highly costly.
To conform with the parsimonious property, we simplify the process by quantizing the video-level temporal span into several sectors; we then predict the likelihood of each relation category's presence within the sectors.
This strategy is distinguished from the segment-based approaches in terms of the feature level because they are not possible to encode the video-level temporal contexts by extracting features from segments.
In contrast, we can utilize the video-level features and predict relations with just a single glance at a video, \ie, there is no need to chunk video into overlapping short segments or repeat computation over the same part.
That is, TSPN can directly specify relationship ranges and categories across the whole video without redundancy.

Let the time span over which two object trajectories intersect (\ie, ${O}_{s} \cap {O}_{o}$) as $\mathcal{T}$.
After $\mathcal{T}$ breaks down into $k$ temporal sectors, one sector is responsible for the range $l$ of the entire time span.
In other words, each temporal sector per pair-of-interest covers a length of $l$, which is calculated as:
\begin{eqnarray}
    l = \frac{1}{k} \left( len \left( \mathcal{O}_{s} \cap \mathcal{O}_{o} \right) \right),
\label{eq:length}
\end{eqnarray}
where $\textit{len}(\cdot)$ stands for the length of a given trajectory.

We can now discretely predict which temporal sectors the relations exist in, instead of directly predicting the temporal span in continuous video space.
We concatenate joint features ${J}_{s}, {J}_{o}, {J}_{u}$ and feed into a fully-connected layer followed by a non-linearity (\eg, sigmoid).
\begin{eqnarray}
    Z = \sigma \left( \textbf{W}_{z}^{\textrm{T}} \left( {J}_{s} \parallel {J}_{o} \parallel {J}_{u} \right) + \textbf{b}_{z} \right),
\label{eq:relation_predict}
\end{eqnarray}
where the output matrix $Z \in \mathbb{R}^{m \times k}$ represents the outer product (denoted as $\otimes$ the equation below) of the probability distribution of relation categories $\mathcal{R} \in \mathbb{R}^{m}$ and that of temporal span $\mathcal{T} \in \mathbb{R}^{k}$, enabling multi-label classification within multiple time spans.
\begin{eqnarray}
    Z = \mathcal{R} \otimes \mathcal{T}.
\label{eq:outer_product}
\end{eqnarray}

\subsection{Loss Function}
\label{sec:3f}
The whole network can be trained in an end-to-end manner. Total loss is sum of relationness ($\mathcal{L}_{R}$) and temporal span ($\mathcal{L}_{T}$) loss. Both are binary cross entropy loss.
\begin{eqnarray}
    \mathcal{L}_{total} = \mathcal{L}_{R} + \mathcal{L}_{T}.
\label{eq:loss}
\end{eqnarray}

$\mathcal{L}_{R}$ measures the loss between $p$ pair of ground truth binary values $r^*$ (pair-of-interest as 1, otherwise 0) and the predicted relationness ($r$ in~\eqnref{eq:relationness}).
\begin{eqnarray}
    \mathcal{L}_{R} = - \frac{1}{p} \sum_{i=1}^{p} r^* \cdot \log r + (1-r^*) \cdot \log (1-r).
\label{eq:loss_r}
\end{eqnarray}

$\mathcal{L}_{T}$ measures the loss between $m \times k$ pair of ground truth temporal spans $t^*$ and predicted temporal spans $t$.
\begin{eqnarray}
    \mathcal{L}_{T} = - \frac{1}{m \times k} \sum_{i}^{m \times k} t^* \cdot \log t + (1-t^*) \cdot \log (1-t).
\label{eq:loss_t}
\end{eqnarray}
We construct a ground truth temporal span into a set of temporal sectors with values of $0$ or $1$, each of which is set to $1$ if the relation lasts more than half within that temporal sector, otherwise set to $0$.

\section{Efficiency of TSPN}
To see how efficient TSPN is, we compare TSPN with two representative VidVRD approaches (\vs segment-based and window-based) in terms of estimated computation time.
The segment-based approach predicts video-level relations by dividing a video into several short-term segments and connecting to longer relations if adjacent segments share the same relations.
The window-based approach first obtains multiple tracklets by sliding varying-size windows over the trajectories, finds tracklet pairs with high similarity via a correlation embedding module, and finally predicts the relations for the valid pairs.
TSPN first finds out what object trajectory pairs are highly likely to have relations by calculating relationness score, then TSPN simultaneously predicts relation categories and their temporal span with sampled pairs.

Note that direct comparison of total computing time is difficult as the greedy association in the segment-based approach is an offline algorithm, and the source codes of the window-based approach are not publicly available.
We thus provide the asymptotic computational complexity of each method for relation prediction.
Here, we do not consider additional features or modules (\eg, greedy association, pair correlation embedding module, and relationness scoring) other than the main components for simplicity.
Also, we consider that the segment length of the segment-based approach, a minimum window size of the window-based approach, and coverage per temporal sector in TSPN are all on the same scale for fair comparison.
Now the time complexity can be represented by the number of object pairs to consider, assuming that predicting relations for each pair of objects takes the same amount of time.

Let the intersecting length of subject and object be $L$, segment or minimum-window or sector length be $l$, and stride be $s$.
Intuitively, the computation time of segment-based approach is proportional to the number of segments.
The number of segments can be calculated as:
\begin{eqnarray}
    \mathbb{N}_{s} = \left\lceil \frac{L-l+s}{s} \right\rceil.
\label{eq:segment-time}
\end{eqnarray}


Similarly, the window-based method depends on the number of windows.
The window-based method exhaustively uses window of all sizes, from smallest (\ie, $l$) to largest (\ie, $L$), to capture both the short-term and long-term object relations.
The number of all windows can be calculated as:
\begin{eqnarray}
    \mathbb{N}_{w} = \sum_{k=1}^{\frac{L}{l}} \left\lceil \frac{L-kl+s}{s} \right\rceil.
\label{eq:window-time}
\end{eqnarray}

Since relation detection requires two windows each responsible for subject and object, the overall computation shows a quadratic growth rate with respect to the number of windows.


In contrast, TSPN directly predicts the relationship categories and their temporal span using the entire video context only once.
Thus, it shows a linear growth rate with respect to the number of temporal sectors.
Since there is no overlap between sectors, the number of sectors can be simplified as:
\begin{eqnarray}
    \mathbb{N}_{t} = \frac{L}{l}.
\label{eq:tspn-time}
\end{eqnarray}

Under the condition of $s < l$, we can obtain the following conclusions\footnote{
Here is a simple derivation: \\
1) $\mathbb{N}_{w}$ can be rewritten as $\mathbb{N}_{s} + \sum_{k=2}^{\frac{L}{l}} \left\lceil \frac{L-kl+s}{s} \right\rceil$; therefore, $\mathbb{N}_{s} < \mathbb{N}_{w}$. \\
2) The lower bound of $\mathbb{N}_{s}$ is $\frac{L-l}{s} + 1$; since $s<l$, it satisfies: $\frac{L-l}{s} + 1 > \frac{L-l}{l} + 1 = \frac{L}{l} = \mathbb{N}_{t}$; therefore, $\mathbb{N}_{t} < \frac{L-l}{s} + 1 \leq \mathbb{N}_{s}$.
}:
\begin{eqnarray}
    \mathbb{N}_{t} \ll \mathbb{N}_{w} < \mathbb{N}_{s},
\label{eq:time-comparison}
\end{eqnarray}
which means that TSPN is always more efficient than the segment and window-based approaches.
In practice, we follow the typical settings of stride/segment length ratio ($s=15$, $l=30$; thus, $s=l/2$), so the estimated computation time of segment and TSPN-based approaches are proportional to $\frac{L}{s}$ and $\frac{L}{l}$, respectively.
Therefore, relation prediction of TSPN is approximately $2{\times}$ more efficient than the segment-based approach and even 4 orders of magnitude efficient than window-based approach.
The overall estimated computation times for relation prediction are summarized in~\tabref{tab:time}.

%% file: 04experiments.tex
\begin{table*}[t!]
    \setlength{\tabcolsep}{8pt}
    \centering
    \resizebox{0.8\linewidth}{!}{
    \begin{tabular}{@{}ccccccccc@{}}
        \toprule
        & & & \multicolumn{3}{c}{\textit{VRDet}} & \multicolumn{3}{c}{\textit{VRTag}} \\ \cmidrule(l){4-6} \cmidrule(l){7-9} 
        \multicolumn{1}{c}{Approach} & \multicolumn{1}{c}{Model} & \multicolumn{1}{c}{Features} & \textit{R@50} & \textit{R@100} & \textit{mAP} & \textit{P@1} & \textit{P@5} & \textit{P@10} \\ \midrule
        \multirow{6}{*}{\textit{Segment}} & \multicolumn{1}{c}{VidVRD~\cite{shang2017video}} & iDT & 5.54 & 6.37 & 8.58 & 43.00 & 28.90 & 20.80 \\
        & \multicolumn{1}{c}{MHRA~\cite{di2019multiple}} & - & 6.82 & 7.39 & 13.27 & 41.00 & 28.70 & 20.95 \\
        & \multicolumn{1}{c}{GSTEG~\cite{tsai2019video}} & iDT+GloVe & 7.05 & 8.67 & 9.52 & 51.50 & 39.50 & 28.23 \\
        & \multicolumn{1}{c}{VRD-GCN~\cite{qian2019video}} & iDT & 8.07 & 9.33 & 16.26 & 57.50 & 41.00 & 28.50 \\
        & \multicolumn{1}{c}{MHA~\cite{su2020video}} & iDT+word2vec & 9.53 & 10.38 & 19.03 & 57.50 & 41.40 & 29.45 \\
        & \multicolumn{1}{c}{VSRN~\cite{cao2021vsrn}} & I3D+word2vec & - & - & - & 43.03 & 33.96 & - \\
        \midrule
        \multirow{1}{*}{\textit{Window}} & \multicolumn{1}{c}{PPN-STGCN~\cite{liu2020beyond}} & I3D+ResNet101 & 11.21 & 13.69 & 18.38 & 60.00 & 43.10 & 32.24 \\
        \midrule
        & \multicolumn{1}{c}{TSPN (Ours)} & ResNet101 & 11.56 & 14.13 & 18.90 & 60.50 & 43.80 & 33.73 \\ \bottomrule
    \end{tabular}
    }
    \caption{
    \textbf{Comparison with recent approaches on ImageNet-VidVRD~\cite{shang2017video} dataset.}
    We list the features used in each model if available.
    They used different features to extract object trajectory representation: improved dense trajectory (iDT) features~\cite{wang2013action}, ResNet101~\cite{he2016deep}, and I3D~\cite{carreira2017quo}.
    Also, some of the methods used language embeddings: word2vec~\cite{mikolov2013distributed} and GloVe~\cite{pennington2014glove}.
    }
\label{tab:vidvrd}
\end{table*}

\begin{table*}[t!]
    \setlength{\tabcolsep}{8pt}
    \centering
    \resizebox{0.8\linewidth}{!}{
    \begin{tabular}{@{}ccccccccc@{}}
        \toprule
        & & & \multicolumn{3}{c}{\textit{VRDet}} & \multicolumn{3}{c}{\textit{VRTag}} \\ \cmidrule(l){4-6} \cmidrule(l){7-9} 
        \multicolumn{1}{c}{Approach} & \multicolumn{1}{c}{Model} & \multicolumn{1}{c}{Features} & \textit{R@50} & \textit{R@100} & \textit{mAP} & \textit{P@1} & \textit{P@5} & \textit{P@10} \\ \midrule
        \multirow{3}{*}{\textit{Segment}} & \multicolumn{1}{c}{\textsc{MMFF}~\cite{sun2019video}} & iDT+word2vec & 6.89 & 8.83 & 6.56 & 51.20 & 40.73 & - \\ 
        & \multicolumn{1}{c}{\textsc{MHA}~\cite{su2020video}} & iDT+word2vec & 6.35 & 8.05 & 6.59 & 50.72 & 41.56 & 32.53 \\
        & \multicolumn{1}{c}{VSRN~\cite{cao2021vsrn}} & I3D+word2vec & - & - & - & 46.15 & 37.00 & - \\
        \midrule
        \multirow{2}{*}{\textit{Window}} & \multicolumn{1}{c}{\textsc{PPN-STGCN}~\cite{liu2020beyond}} & I3D+ResNet101 & 8.21 & 9.90 & 6.85 & 48.92 & 36.78 & - \\
        & \multicolumn{1}{c}{Gao~\etal~\cite{gao2021video}} & ResNet101+GloVe & 8.35 & 10.21 & 9.33 & - & 52.73 & 41.14 \\
        \midrule
        & \multicolumn{1}{c}{TSPN (Ours)} & ResNet101 & 9.33 & 10.71 & 7.61 & 53.14 & 42.22 & 34.94 \\ \bottomrule
    \end{tabular}
    }
    \caption{\textbf{Comparison with recent approaches on VidOR~\cite{shang2019annotating} validation dataset.}}
\label{tab:vidor}
\end{table*}

\section{Experiments}
\label{sec:experiments}
In this section, we conduct comprehensive studies to validate the effectiveness of the proposed TSPN. 
Here, we report results on two challenging datasets: ImageNet-VidVRD~\cite{shang2017video} and VidOR~\cite{shang2019annotating}.
To understand the behavior of TSPN, we provide extensive quantitative analyses.
Lastly, we present qualitative results to examine how our model can benefit from temporal globality concretely.

\subsection{Implementation Details}
Following~\cite{liu2020beyond}, we adopt Faster R-CNN~\cite{ren2015faster} and DeepSORT~\cite{wojke2017simple} for object trajectory detection. 
We first generate bounding boxes at each frame using Faster R-CNN~\cite{ren2015faster} equipped with ResNet-101~\cite{he2016deep} backbone.
The Faster R-CNN is pre-trained on MS-COCO~\cite{lin2014microsoft} and ILSVRC2016-DET~\cite{russakovsky2015imagenet} datasets with $35$ object categories.
We perform NMS in the detection phase with an IoU threshold of $0.5$ to eliminate redundant bounding boxes since NMS in the tracking phase may wrongly remove overlapping trajectories with different classes because it is class-agnostic.
As a multi object tracker, we adopt DeepSORT~\cite{wojke2017simple}, which integrates visual appearance information to gain robustness against identity switching while associating the objects with the same identity in neighboring frames.
We extract fixed-size features of video using ResNet 101 backbone~\cite{he2016deep} and RoI Align operation~\cite{he2017mask} to get fixed-length representations from varying-size object proposals.
For TSPN training, we use Adam optimizer~\cite{kingma2014adam} with a learning rate of $10^{-3}$ and weight decay of $10^{-4}$ for a batch size of 32.
We set the number of pair-of-interests as $64$ (\ie, $p=64$), and the number of sectors is set to $16$ (\ie, $k=16$).
We keep top-$100$ relation triplet predictions per video.
The hyperparameter configurations are set to be the same on both ImageNet-VidVRD~\cite{shang2017video} and VidOR~\cite{shang2019annotating} datasets.

\subsection{Dataset Configuration}
In both datasets, objects and relation triplets (with temporal span) are annotated in the form of $\{$\textit{trajectory id, category, bounding box coordinates}$\}$ and $\{$\textit{(subject id -- predicate category -- object id), (start frame -- end frame)}$\}$, respectively.
We exclude several unannotated videos in the experiment.

\paragraph{ImageNet-VidVRD}
ImageNet-VidVRD~\cite{shang2017video} is a subset of ILSVRC2016-VID~\cite{russakovsky2015imagenet} train and validation set. Videos are selected with the criteria of whether they
contain clear visual relations. It contains $1{,}000$ videos (train:test split is $800:200$) with the manually labeled object categories, corresponding bounding box trajectories, and relation triplets.
It covers $35$ and $132$ categories for objects and predicates, respectively.
Note that the relations in the training set are all annotated in the segment-level (\eg, $30$ frames). Thus, we link them into the video-level relations in order to directly optimize long-term relations.

\paragraph{VidOR}
VidOR~\cite{shang2019annotating} is a large-scale video dataset which contains $10{,}000$ videos (train:val:test split is $7{,}000:835:2{,}165$) from YFCC100M~\cite{thomee2016yfcc100m} collection. 
The average video length in train/val set is ${\sim}36$ seconds, totaling ${\sim}84$ hours.
It contains $80$ categories of objects with trajectories to indicate their spatio-temporal location in the videos, $50$ categories of relation predicates, and carefully annotated the relation triplets.
This results in around $50{,}000$ objects and $380{,}000$ relation instances annotated.
Note that only the training and validation set are used for the experiment because the test set is not yet publicly available.

\subsection{Evaluation Protocol}
We follow the standard evaluation task settings and metrics~\cite{shang2017video} listed below:
\paragraph{Tasks}
In \textit{Visual Relation Detection (VRDet)} task, given a video, the objective is to jointly predict object trajectories, categories, and existing relations with its duration. The object trajectories are considered matched only if vIoU\footnote{vIoU
denotes the volumetric intersection over the union of two tubes.} with the ground truth trajectory is greater than half.

In \textit{Visual Relation Tagging (VRTag)} task, given video with a set of ground truth object trajectories, the objective is to find all object categories and existing relations. In other words, it needs to detect the relation triplet $<subject-predicate-object>$ correctly.

In both tasks, detected visual relation instances are treated as correct only if the trajectories of subject and object that form the relation both have sufficiently high vIoU (\ie, vIoU $>$ 0.5) with the ground truth.

\paragraph{Evaluation metrics}
For \textit{Visual Relation Detection (VRDet)} task, we adopt mean Average Precision ($mAP$) and $Recall@K$ to evaluate the detection performance. Specifically, we use video-wise $Recall@50$ and $Recall@100$, which measures the fraction of the top-$K$ predictions among the ground-truth triplets.
Following~\cite{shang2017video}, we also report the $mAP$ metric to evaluate the overall precision performance at different recall values.

For \textit{Visual Relation Tagging (VRTag)} task, $Precision@K$ as the evaluation metric to emphasize the ability to tag accurate visual relations. It measures the fraction of ground-truth among the top-$K$ triplets. Specifically, we use video-wise $Precision@1$, $Precision@5$, and $Precision@10$ since the number of visual relation instances per segment is $9.5$.

\subsection{Comparison with State-of-the-Art}
We compare TSPN with state-of-the-art approaches on the two VidVRD benchmarks: ImageNet-VidVRD~\cite{shang2017video} and VidOR~\cite{shang2019annotating}.
For a comparison, we consider the following methods: VidVRD~\cite{shang2017video}, MHRA~\cite{di2019multiple}, GSTEG~\cite{tsai2019video}, VRD-GCN~\cite{qian2019video}, MMFF~\cite{sun2019video}, MHA~\cite{su2020video}, VSRN~\cite{cao2021vsrn}, PPN-STGCN~\cite{liu2020beyond}, and gao \etal~\cite{gao2021video}.
Existing VidVRD approaches can be divided into two families: segment-based~\cite{shang2017video,di2019multiple,tsai2019video,qian2019video,su2020video,cao2021vsrn} and window-based~\cite{liu2020beyond,gao2021video} approaches.
They adopt different methods to generate object trajectory representations including improved dense trajectory (iDT) features~\cite{wang2013action}, ResNet101~\cite{he2016deep}, and I3D~\cite{carreira2017quo}.
Also, some of them used language embeddings such as word2vec~\cite{mikolov2013distributed} or GloVe~\cite{pennington2014glove}.
The results are presented in~\tabref{tab:vidvrd} and~\tabref{tab:vidor}.
In general, the VidOR dataset has longer video lengths than the ImageNet-VidVRD dataset, hence the performance tends to be lower in the same metric.
Overall, we observe that TSPN achieves competitive results on both ImageNet-VidVRD and VidOR benchmarks, demonstrating its effectiveness while having lower computational cost than counterparts.
In ImageNet-VidVRD benchmark, when evaluated with \textit{VRDet} mAP metric, MHA~\cite{su2020video} slightly performs better ($+$0.13) than TSPN, albeit our TSPN outperforms all methods in all other metrics without the need of any external resources (\eg, language embeddings).
In VidOR benchmark, our TSPN mostly outperforms its counterparts and performs on par with Gao~\etal~\cite{gao2021video}.
Generally, our TSPN performs better than Gao~\etal~\cite{gao2021video} in \textit{VRDet} task which requires both object and relation detection at the same time.
TSPN leverages the complementariness of multiple information (visual, geometric, and semantic), and is designed to utilize the spatio-temporal contexts of the entire video, making it strong not only for the short-term but also for the long-term relations.

\subsection{Quantitative Analysis}
All analytic experiments are conducted on the ImageNet-VidVRD dataset.

\begin{table}[t!]
    \setlength{\tabcolsep}{6pt}
    \centering
    \scriptsize
    \resizebox{\linewidth}{!}{
    \begin{tabular}{@{}cccccccccc@{}}
        \toprule
        & \multicolumn{3}{c}{Features} &  \multicolumn{3}{c}{\textit{VRDet}} & \multicolumn{3}{c}{\textit{VRTag}} \\ \cmidrule(l){2-4} \cmidrule(l){5-7} \cmidrule(l){8-10}
        Exp &    $A$   &    $B$   &    $C$   & \textit{R@50}  & \textit{R@100} & \textit{mAP} & \textit{P@1} & \textit{P@5} & \textit{P@10}   \\ \midrule
        1   &\checkmark&          &          &  8.84  &  12.24  & 15.49 & 54.50  & 39.70  & 29.65   \\
        2   &\checkmark&\checkmark&          &  9.85  &  13.13  & 16.94 & 57.00  & 41.90  & 31.25   \\
        3   &\checkmark&          &\checkmark&  10.94 &  13.84  & 18.33 & 59.00  & 43.00  & 32.85   \\
        \rowcolor{Gray}
        4   &\checkmark&\checkmark&\checkmark& \textbf{11.56} & \textbf{14.13} & \textbf{18.90} & \textbf{60.50} & \textbf{43.80} & \textbf{33.73} \\ \bottomrule
    \end{tabular}
    }
    \caption{
        \textbf{Effect of feature combination in Relationness Scoring module.}
        $A$ refers to visual, $B$ to geometric, and $C$ to semantic features.
    }
\label{tab:rel_scoring}
\end{table}

\begin{table}[t!]
    \setlength{\tabcolsep}{7pt}
    \centering
    \scriptsize
    \resizebox{\linewidth}{!}{
    \begin{tabular}{@{}ccccccccc@{}}
        \toprule
        & \multicolumn{2}{c}{Ablations} &  \multicolumn{3}{c}{\textit{VRDet}} & \multicolumn{3}{c}{\textit{VRTag}} \\ \cmidrule(l){2-3} \cmidrule(l){4-6} \cmidrule(l){7-9}
        Exp &  $\mathcal{R}$      &  $\mathcal{T}$      & \textit{R@50}  & \textit{R@100} & \textit{mAP} & \textit{P@1} & \textit{P@5} & \textit{P@10}   \\ \midrule
        1   &          &          &  6.14  &  7.52  &  9.38 & 40.50  & 28.90  & 21.95   \\
        2   &\checkmark&          &  9.62  &  11.69  & 16.29 & 57.00  & 41.90  & 30.45   \\
        3   &          &\checkmark&  9.95  &  12.35  & 15.93 & 54.50  & 40.20  & 29.50   \\
        \rowcolor{Gray}
        4   &\checkmark&\checkmark& \textbf{11.56} & \textbf{14.13} & \textbf{18.90} & \textbf{60.50} & \textbf{43.80} & \textbf{33.73} \\ \bottomrule
    \end{tabular}
    }
    \caption{
        \textbf{Model ablations.} Here, $\mathcal{R}$ and $\mathcal{T}$ denote relationness scoring module and temporal span proposal module, respectively.
    }
\label{tab:ablation}
\end{table}

\paragraph{Contribution of different information for relationness scoring}
In~\tabref{tab:rel_scoring}, we ablate the feature combinations in relationness scoring module.
We used RoI appearance features ($A$), bounding box coordinates ($B$), and classification probability distribution ($C$) as visual, geometric, and semantic information for relationness scoring.
By default, we only used visual information (Exp 1) to measure relationness score.
We then progressively fuse geometrical information (Exp 2) and semantic information (Exp 3).
The bounding box coordinates of object trajectories ($B$) can provide cues about the spatial relations (\eg, next to, in front of, \textit{etc}.), and the classification probability distribution ($C$) can be used to leverage inter-object bias.
It is known that inter-object bias can help determining relationship between them~\cite{zellers2018neural}.
When all information are combined (Exp 4), the model achieves the best performance in both \textit{VRDet} and \textit{VRTag} tasks, implying that each information contributes to the performance in different aspect.
From the results, we confirm the importance of the complementary feature fusion.

\paragraph{TSPN Ablations}
We consider several ablations on building blocks of TSPN to identify how each component contributes to the performance and verify its efficacy.
Ablation results are summarized in~\tabref{tab:ablation}.
Without both relationness scoring ($\mathcal{R}$) and temporal span proposal ($\mathcal{T}$) (equivalent to Exp 1), we have to optimize ${m} \times {l}$ distribution for ${n}^{2}$ pairs, but with both $\mathcal{R}$ and $\mathcal{T}$ (equivalent to Exp 4), we only need to optimize ${m} \times {k}$ distribution for $p$ pairs, where $p \ll {n}^{2}$ and ${k} \ll {l}$.
In this experiment, we can see that the $\mathcal{R}$ and $\mathcal{T}$ are complementary to each other in constructing a strong VidVRD model.

\begin{table}[t!]
    \setlength{\tabcolsep}{6pt}
    \centering
    \scriptsize
    \resizebox{\linewidth}{!}{
    \begin{tabular}{@{}cccccccc@{}}
        \toprule
            &  &  \multicolumn{3}{c}{\textit{VRDet}} & \multicolumn{3}{c}{\textit{VRTag}} \\ \cmidrule(l){3-5} \cmidrule(l){6-8}
        Exp &  \multicolumn{1}{c}{\# of PoIs ($p$)} & \textit{R@50}  & \textit{R@100} & \textit{mAP} & \textit{P@1} & \textit{P@5} & \textit{P@10}   \\ \midrule
        1   &  16   &  9.85  & 11.84  & 18.03  & \textbf{62.00} & \textbf{44.60}  & \textbf{34.96} \\
        2   &  32  & 10.32  & 12.93  & 18.41  & 61.50 & 44.10  & 34.04 \\
        \rowcolor{Gray}
        3   &  \textbf{64}  & \textbf{11.56} & \textbf{14.13} & \textbf{18.90} & 60.50 & 43.80 & 33.73 \\
        4   &  128  & 10.92  & 14.29  & 18.29  & 57.50 & 43.10  & 32.68 \\ \bottomrule
    \end{tabular}
    }
    \caption{\textbf{Optimal number of Pair-of-Interests (PoIs).}}
\label{tab:optimal_poi}
\end{table}

\paragraph{Optimal Size of Relation Search Space}
We further explore the optimal number of pair-of-interest ($p$) per video.
Considering that VidVRD contains $4{,}835$ video-level relation instances in 200 test set (${\sim}24$ instances per video on average), we set the number of pair-of-interest as $32$ by default.
We examine several variations ($p=16$ or $64$ or $128$) in~\tabref{tab:optimal_poi}. Although 16 shows the highest precision since it has a low false-positive rate, $64$ holds the best overall results.
The results indicate that balancing positives and negatives labels is crucial since the TSPN learns from penalizing the false positives.
The number of PoIs should be appropriately set because the resulting sampled object pairs affect the subsequent processing.
If it is too large (\eg, $128$), the probability of the noisy object pairs being included in the PoIs becomes higher and result in degraded performance.
We empirically found that $64$ works well, and use this number for the rest of the experiments.

\begin{table}[t!]
    \setlength{\tabcolsep}{5pt}
    \centering
    \scriptsize
    \resizebox{\linewidth}{!}{
    \begin{tabular}{@{}cccccccc@{}}
        \toprule
            &  &  \multicolumn{3}{c}{\textit{VRDet}} & \multicolumn{3}{c}{\textit{VRTag}} \\ \cmidrule(l){3-5} \cmidrule(l){6-8}
        Exp &  \multicolumn{1}{c}{\# of sectors (${k}$)} & \textit{R@50}  & \textit{R@100} & \textit{mAP} & \textit{P@1} & \textit{P@5} & \textit{P@10}   \\ \midrule
        1   &  4   &  5.19  &  6.25  & 9.79  & 33.50 & 23.10  & 15.44 \\
        2   &  8  &  10.02  & 11.49  & 14.74  & 49.00 & 36.50  & 24.04 \\
        \rowcolor{Gray}
        3   &  \textbf{16}  & \textbf{11.56} & 14.13 & \textbf{18.90} & \textbf{60.50} & \textbf{43.80} & \textbf{33.73} \\
        4   &  32  & 11.32  & \textbf{14.50}  & 18.69  & 59.50 & 42.60  & 33.14 \\ \bottomrule
    \end{tabular}
    }
    \caption{
    \textbf{Effect of the number of temporal sectors.}
    }
\label{tab:quantization}
\end{table}

\begin{figure*}[t!]
    \centering
    \includegraphics[width=\linewidth]{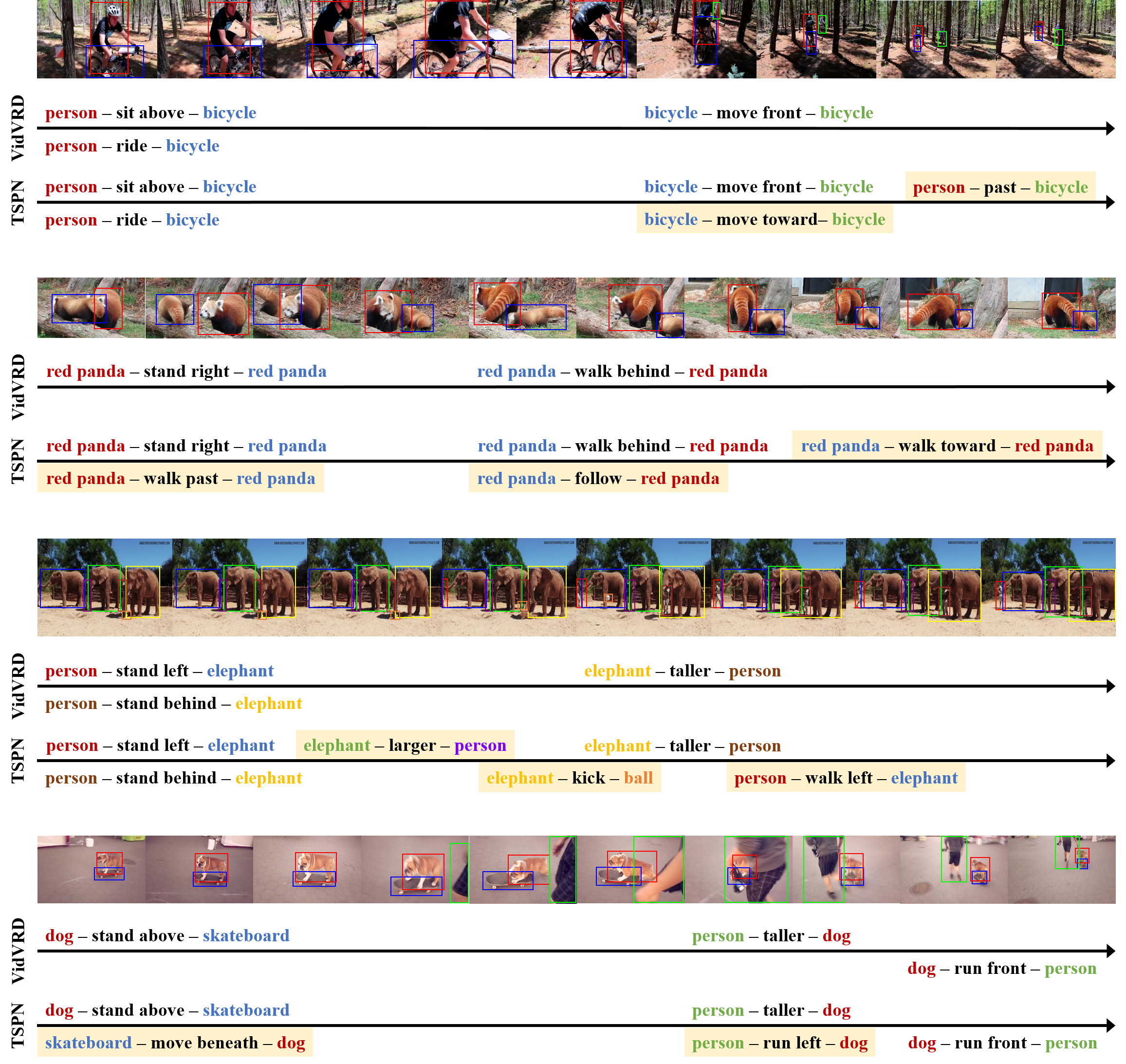}
    \caption{
        \textbf{Qualitative examples of visual relation detection results.} For comparison, we contrast the predicted relation triplets (\ie, \texttt{subject-relation-object}) of VidVRD with those of TSPN for each given video.
        The same color means the same object instance.
        The arrows represent the time axes, providing an approximation of the temporal span of the predicted relation triplets.
        We highlight relations that TSPN correctly predicted, while VidVRD did not.
        The predicted relations are considered correct only if the pair of object trajectories have sufficiently high vIoU (\ie, vIoU $>$ 0.5) with ground truth trajectories, and only the correct relations of the top-20 predictions are shown in the figure.
    }
\label{fig:qualitative}
\end{figure*}

\paragraph{Trade-off in Temporal Span Quantization}
We investigate the effect of quantization of temporal spans by varying the number of sectors.
We divide the intersecting temporal span of subject and object trajectories into $k$ sectors ($k$ of \eqnref{eq:length}).
Since the intersection of each object trajectory pair varies, each object pair may have a different quantization impact.
Sparse quantization narrows the search space thus easy to optimize but lowers the matching rate at the same time.
On the other hand, dense quantization increases the matching rate, but it also broadens the search space thus expensive to optimize.
We conduct an experiment in~\tabref{tab:quantization} to assess the optimal degree of quantization.
A limited number of sectors ($k=4, 8$) exhibit sub-optimal performances.
On the other hand, the model achieves good performances when videos are divided into an adequate number of sectors ($k=16, 32$).
Between $k=16$ and $k=32$, we observe that the model performs slightly better when $k$ is set to $16$ than when $k$ is set to $32$.

\subsection{Qualitative Analysis}
To better see how TSPN understands the dynamics of spatio-temrpoal interactions, we provide qualitative examples in~\figref{fig:qualitative}.
Here, we compare TSPN with VidVRD~\cite{shang2017video}, which first proposed a popular segment-based approach.
The results show that the VidVRD often fails to capture long-term relations since it relies on temporarily local features.
On the other hand, TSPN successfully identifies them by leveraging global video context.
For example, the relation \texttt{person-past-bicycle} ($105{-}150$)\footnote{The numbers in parentheses mean ground truth ``begin frame -- end frame".} in the first row can only be detected by understanding the long-term relationship since the model should capture the entire process of \texttt{person} being behind \texttt{bicycle} and catching up with \texttt{bicycle} as it moves forward.
It goes the same for the relation \texttt{red panda-walk past-red panda} ($0{-}75$) in the second row.
TSPN also detects more active or sophisticated relations such as \texttt{elephant-kick-ball} ($0{-}60$) in the third row and \texttt{skateboard-move beneath-dog} ($0{-}360$) in the fourth row, while VidVRD simply detects trivial relations (\eg, \texttt{elephant-taller-person}, \texttt{person-taller-dog}), which can also be predicted in a still image.
This also reveals the limitation of segment-based methods.
That is, they cannot leverage the video-level context.
VidVRD is unable to employ temporarily global features directly since it predicts segment-level relations and aggregates them into video-level relations.
In contrast, TSPN are designed to utilize video-level context to predict long-term and short-term relations from the proposed temporal span.

%% file: 05conclusion.tex
\section{Conclusion}
This work introduces a novel Temporal Span Proposal Network (TSPN) for Video Visual Relation Detection (VidVRD).
TSPN guides what and when to look in order to detect relationships, making it efficient yet effective.
TSPN effectively reduces the relational search space by learning which object pairs should be considered based on the relationness score between object trajectory pairs (relationness scoring module).
TSPN simultaneously predicts the temporal span and categories of the entire relations with a single global video feature, which is not only efficient but also effective in predicting both short and long-term relations (relation \& temporal span prediction module).
We validate the TSPN through comprehensive experiments.
TSPN establishes a strong baseline for VidVRD by leveraging the complementariness of the two key modules.
In particular, TSPN achieves state-of-the-art performance on two VidVRD benchmarks (ImageNet-VidVRD and VidOR) without many external algorithms or heuristics (\eg, greedy association, window sliding) commonly seen in previous approaches while $2{\times}$ or more faster than conventional approaches (segment-based and window-based approaches).

%% file: 08references.tex
\ifCLASSOPTIONcaptionsoff
  \newpage
\fi



{\footnotesize 
\bibliographystyle{IEEEtran}
\bibliography{ieee_tnnls}
}
%

%% file: 09biography.tex
%

\begin{IEEEbiography}[{\includegraphics[width=1in,height=1.25in,clip,keepaspectratio]{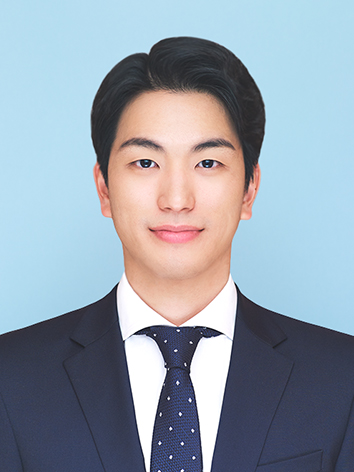}}]{Sangmin Woo} (Student Member, IEEE)
    is currently pursuing the Ph.D. degree in electrical engineering at Korea Advanced Institute of Science and Technology (KAIST), Daejeon, Korea. He received an M.S. degree in Electrical Engineering and Computer Science from Gwangju Institute of Science and Technology (GIST), Gwangju, Korea, in 2021, and a B.S. degree in Electrical Engineering from Kyungpook National University, Daegu, Korea, in 2019. His research interests lie in computer vision and machine learning, especially in a high-level visual understanding.
\end{IEEEbiography}

\begin{IEEEbiography}[{\includegraphics[width=1in,height=1.25in,clip,keepaspectratio]{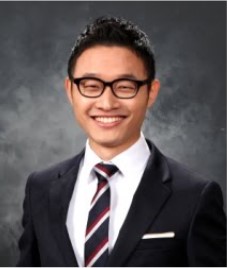}}]{Junhyug Noh} (Member, IEEE)
    is a postdoctoral researcher at Lawrence Livermore National Laboratory (LLNL). He received the B.S. in Computer Science and Engineering \& Statistics from Seoul National University in 2013, and the M.S. and Ph.D. in Computer Science Engineering from Seoul National University in 2015 and 2020, respectively. His research has focused on artificial intelligence, machine learning, and computer vision with a particular interest in object detection and its related high-level vision tasks such as semantic/instance segmentation, scene understanding, and image captioning.
\end{IEEEbiography}


\begin{IEEEbiography}[{\includegraphics[width=1in,height=1.25in,clip,keepaspectratio]{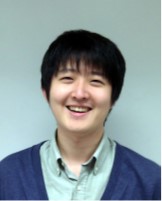}}]{Kangil Kim} (Member, IEEE)
    received the B.S. degree in computer science from the Korea Advanced Institute of Science and Technology, Daejeon, South Korea, in 2006, and the Ph.D. degree from Seoul National University, Seoul, South Korea, in 2012. He was a Senior Researcher with the Natural Language Processing Group, Electronics and Telecommunications Research Institute, Seoul, until 2016, and an Assistant Professor with the Computer Science and Engineering Department, Konkuk University, until 2019. He is currently an Assistant Professor with the Electronics Engineering and Computer Science Department and Artificial Intelligence Graduate School, Gwangju Institute of Science and Technology, Gwangju, South Korea. His research interests include artificial intelligence, evolutionary computation, machine learning, and natural language processing.
\end{IEEEbiography}


